\documentclass[acmsmall]{acmart}
\AtBeginDocument{%
  }

\setcopyright{acmlicensed}
\copyrightyear{2025}
\acmYear{2025}
\acmDOI{XXXXXXX.XXXXXXX}

\acmJournal{JACM}
\acmVolume{37}
\acmNumber{4}
\acmArticle{131}
\acmMonth{8}

\usepackage{times}
\usepackage{soul}
\usepackage{url}
\usepackage{color}
\usepackage{amsmath}
\usepackage{booktabs}
\usepackage{multirow}
\usepackage{array}
\usepackage{caption} 
\usepackage{makecell} 
\usepackage{subcaption}

\begin{document}

\title{MLlm-DR: Towards Explainable Depression Recognition with MultiModal Large Language Models}

\author{Wei Zhang}
\email{zhangwei23@nudt.edu.cn}
\orcid{0009-0009-8079-3349}
\affiliation{
  \institution{National University of Defense Technology}
  \city{ChangSha}
  \country{China}}
  
\author{Juan Chen}
\email{chenjuan@ict.ac.cn}
\affiliation{
  \institution{University of Chinese Academy of Sciences}
  \city{BeiJing}
  \country{China}}

\author{En Zhu}
\authornotemark[1]
\affiliation{
  \institution{National University of Defense Technology}
  \city{ChangSha}
  \country{China}}
\email{enzhu@nudt.edu.cn}
  
\author{Wenhong Cheng}
\affiliation{
  \institution{Shanghai Mental Health Center, Shanghai Jiao Tong University School of Medicine}
  \city{shanghai}
  \country{China}}
\email{chengwhb@aliyun.com}

\author{YunPeng Li}
\affiliation{
  \institution{Nanjing Industria Tenebris Information Technology Co., Ltd.}
  \city{NanJing}
  \country{China}}
\email{yunpli@itenebris.com}

\author{Yuhan Li}
\affiliation{
  \institution{National University of Defense Technology}
  \city{ChangSha}
  \country{China}}
\email{liyuhan@nudt.edu.cn}

\author{Yanbo J. Wang}
\authornote{Corresponding author}
\affiliation{
  \institution{National University of Uzbekistan named after Mirzo Ulugbek}
  \city{Tashkent}
  \country{Uzbekistan}}
\email{wangyanbo@lyzdfintech.com}

\begin{abstract}

Clinical depression diagnosis relies heavily on both verbal and non-verbal cues in patient interviews, yet existing automated methods often operate as black-box models and fail to provide trustworthy explanations, limiting their clinical applicability. Moreover, depression datasets commonly impose strict privacy constraints that prohibit access to raw audio–visual data, 
and the deployment of large LLMs in real-world medical environments is often constrained by computational cost limitations.
To address these challenges, we propose MLlm-DR, a multimodal large language model for explainable depression recognition. We first employ knowledge distillation to transfer high-quality diagnostic rationales from a powerful LLMs to a smaller, deployable LLMs, thereby equipping it with clinically aligned reasoning abilities. We further introduce a lightweight query module (LQ-former) that extracts salient depression-related cues from pre-extracted audio and visual features and maps them into an LLM-compatible representation. MLlm-DR is trained in two stages: multimodal alignment via LQ-former pretraining, followed by multi-task optimization that jointly learns rationale generation and score regression, encouraging consistency and improving interpretability. 
Experimental results show that MLlm-DR achieves state-of-the-art performance on two interview-based benchmark datasets, CMDC and E-DAIC-WOZ, while simultaneously generating clinically meaningful and readable explanations.

\end{abstract}




\maketitle

\section{Introduction}

Clinical interviews are the cornerstone of depression diagnosis, widely regarded as the gold standard \cite{gold}. During these interviews, clinicians assess symptom severity based on established criteria like DSM-5, ICD-11, and CCMD-3 \cite{DSM-5,ICD-11,CCMD-3}, evaluating indicators such as suicidal ideation, anhedonia, and sleep disturbances. The final diagnosis is derived from a holistic evaluation of these symptoms. However, this manual process is not only time-consuming and resource-intensive but also susceptible to the subjective judgment of individual clinicians, leading to potential inconsistencies.

To mitigate these limitations, automated depression diagnosis has emerged as a promising research avenue \cite{sub-attentional,MMFF,End-to-end}. Early approaches successfully analyzed dialogue semantics and extracted affective cues from speech and facial expressions, achieving high diagnostic efficiency and accuracy \cite{iifdd,multimodal,MDDR,mago}. Yet, a critical flaw persists in these methods: their "black-box" nature. While they can predict a depression score, they fail to provide a rationale for their predictions. This lack of explainability is a major barrier to clinical adoption, as practitioners are hesitant to trust a diagnostic tool without a clear, verifiable reasoning process.

The advent of Large Language Models (LLMs) \cite{qwen2.5,llama,gpt4,ernie,glm} offers a compelling solution to this explainability crisis. With their profound capabilities in understanding context \cite{zhaochat,Instructerc,fado,huang}, performing logical reasoning \cite{distilling,chain,kojima}, and generating coherent text \cite{T5,qwen2.5}, LLMs can not only assess a patient's depression level from interview transcripts but also articulate a detailed, evidence-based rationale for their conclusion. This ability to mirror the clinician's process of "scoring with explanation" promises to significantly enhance the trustworthiness and clinical utility of automated systems.

However, a purely text-based approach is fundamentally incomplete. Depression manifests not only in the content of a patient's speech but also in their non-verbal behaviors. Cues such as vocal prosody (e.g., flat intonation) and facial expressions (e.g., lack of affect) are indispensable diagnostic indicators \cite{tmac,facial} that provide a more objective and holistic view of a patient's mental state. While state-of-the-art multimodal LLMs (MLLMs) like Gemini and Claude 3 can process raw video data \cite{2023gemini,2024claude,Qwen2.5-VL}, they are impractical for real-world clinical applications. Due to stringent patient privacy regulations and ethical considerations, depression datasets almost never provide raw videos, but rather pre-extracted, anonymized features. This data modality gap renders existing end-to-end MLLMs unusable, presenting a significant technical challenge: \textbf{how can we equip an LLMs with multimodal understanding capabilities when direct access to the original modalities is prohibited?}

Furthermore, deploying frontier LLMs in clinical settings is often infeasible due to their massive computational footprint and operational costs. A practical solution demands the use of smaller, more efficient models. This introduces a second, equally critical challenge: \textbf{how can a small-scale LLMs achieve the sophisticated reasoning and explanation-generation abilities typically associated with its large-scale counterparts, especially in a specialized domain like clinical psychology?}

\begin{figure}[!h]
\centering
\includegraphics[width=0.7\linewidth]{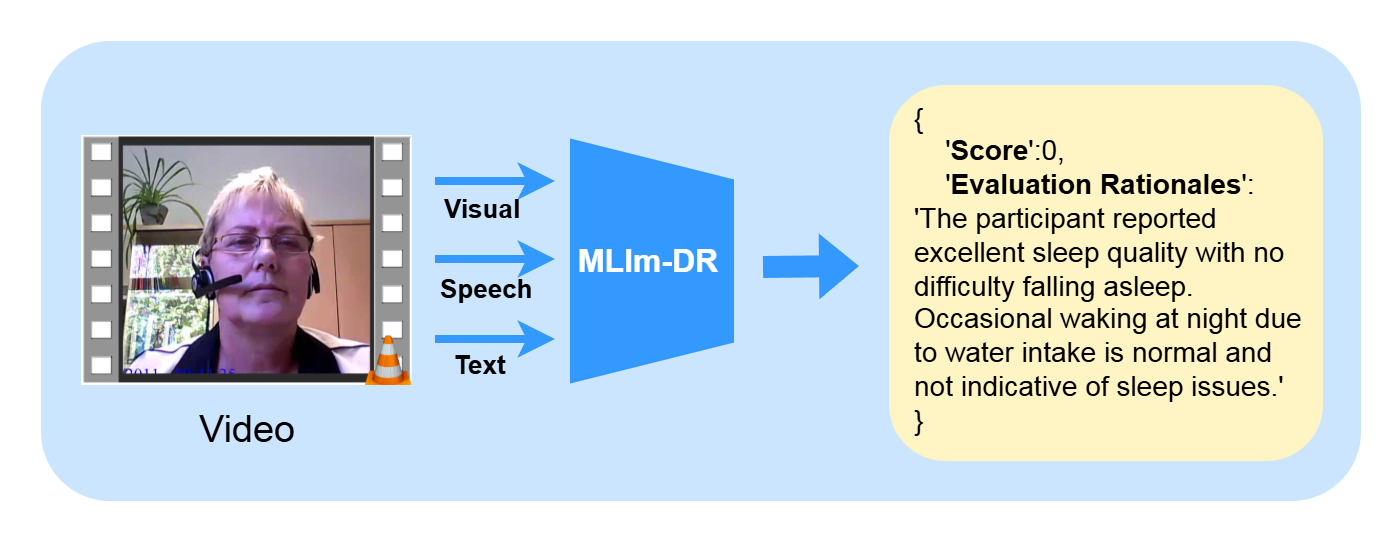}
\caption{
The Multimodal Large Language Model (MLlm-DR) is designed for explainable depression recognition. It leverages transcribed text, speech, and visual data from participants' interview videos to generate depression scores and corresponding evaluation rationales. }
\label{fig_1}
\end{figure}

To address these intertwined challenges of modality alignment under privacy constraints and reasoning distillation for model practicality, we propose \textbf{MLlm-DR}, a novel Multimodal Large Language Model for explainable Depression Rationale generation. As illustrated in Figure~\ref{fig_1}, our framework introduces a solution that is both effective and practical. At its core, MLlm-DR is built upon a small-scale LLMs to ensure deployment feasibility. To imbue it with expert-level reasoning, we introduce a knowledge distillation strategy: we leverage a powerful, large-scale LLMs to generate high-quality diagnostic rationales from interview transcripts, creating a rich, explainable training corpus. Fine-tuning our small model on this distilled knowledge endows it with robust, clinically-aligned reasoning capabilities.

To tackle the multimodal challenge, we design a novel and lightweight projection module, the Lightweight Query-based Transformer (\textbf{LQ-former}). Instead of processing raw data, LQ-former operates on pre-extracted audio and visual features, respecting data privacy. It employs a small set of learnable query vectors to dynamically probe the feature sequences and distill the most salient depression-related signals. These signals are then projected into a compact sequence of embeddings, effectively translating non-verbal cues into a representation the LLMs can process. We employ a principled, two-stage training strategy. First, we train only the LQ-former, teaching it to align multimodal features with the frozen LLM's latent space. This decouples modality alignment from task-specific learning.

Second, with the trained LQ-former frozen, we fine-tune the LLMs using a synergistic multi-task objective. This objective jointly optimizes a cross-entropy loss for rationale generation and a Mean Squared Error (MSE) loss for the final score prediction. This dual-task optimization creates a powerful feedback loop: it compels the model to generate explanations that are not only textually coherent but also quantitatively consistent with the predicted depression score. This reciprocal constraint acts as a form of regularization, ensuring the generated rationales are more accurate and clinically grounded, thereby enhancing the model's overall diagnostic fidelity.

Our contributions are summarized as follows:
\begin{itemize}
    \item We propose MLlm-DR, the first framework, to our knowledge, that enables a \textbf{small-scale LLMs to perform explainable, multimodal depression diagnosis}, addressing the key practical challenges of model efficiency and restricted data access.
    
    \item We introduce a \textbf{knowledge distillation strategy} that effectively transfers the complex reasoning capabilities of a large-scale LLMs to a compact, efficient model, enabling it to generate high-quality, clinically-aligned diagnostic rationales.
    
    \item We design a novel, lightweight projection module, \textbf{LQ-former}, that efficiently aligns pre-extracted audio-visual features with the LLM's latent space, enabling effective multimodal understanding \textbf{without requiring access to raw data}.
    
    \item Extensive experiments on the CMDC and E-DAIC-WOZ datasets demonstrate that our method achieves \textbf{state-of-the-art performance}, validating the effectiveness of our proposed framework in generating accurate and interpretable diagnostic results.
\end{itemize}

\section{RELATED WORK}

\subsection{Multimodal Depression Recognition}

Recently, multimodal fusion methods have made valuable progress in many depression recognition tasks. 
For instance, Wei et al. \cite{sub-attentional} design independent attention fusion modules for each PHQ-8 sub-score to extract multimodal features relevant to specific sub-scores, generating individual sub-scores and ultimately aggregating them for overall depression evaluation. 
Yuan et al. \cite{MMFF} introduce multi-order factor decomposition to extract features from single modalities and their cross-modal combinations, significantly enhancing the representational capacity of multimodal learning and improving model interpretability through a dynamic weighting mechanism.
Jung et al. \cite{HiQuE} explicitly model the hierarchical structure of interview questions (primary questions and follow-ups), simulating the diagnostic logic of clinicians while leveraging attention mechanisms to identify critical questions and features, enabling more precise depression detection.
However, these methods rely on neural networks to directly predict depression scores without providing the corresponding rationale, which reveals a significant limitation.

\subsection{LLMs-based Dialogue Understanding}

Recently, large language models (LLMs) have made significant progress in dialogue understanding tasks. 
For instance, 
Li et al. \cite{empirical} demonstrate the potential of LLMs in Dialogue Relation Extraction (DRE) tasks, showing their superior ability to capture long-span and multi-turn dialogue information, outperforming traditional sequence-based and graph-based methods. 
Lei et al. \cite{Instructerc}  investigate the use of LLMs for Emotion Recognition in Conversations (ERC). They pretrain the model on a speaker identification task to capture emotional expression characteristics of different roles in dialogues and fine-tune it through multi-task learning by integrating ERC and emotion influence prediction tasks, enhancing performance.
Additionally, Huang et al. \cite{huang} propose the Emotion-Cause Reasoning Chain (ECR-Chain) framework, leveraging LLMs to analyze statements in dialogues that trigger target emotions and thereby predict causal emotion entailment (CEE). This framework incorporates cognitive appraisal theory to deeply explore the process of emotion generation, thus providing strong interpretability for reasoning outcomes. 
These studies collectively demonstrate the tremendous potential of LLMs in advancing dialogue understanding tasks.

\subsection{Cross-Modal Semantic Alignment}

To enable LLMs to process and understand multimodal inputs, researchers have proposed various cross-modal semantic alignment methods\cite{flamingo,cogvlm}, achieving significant progress in multiple cross-modal language generation tasks, such as visual question answering and image captioning.
For example, Li et al. \cite{Q} propose a semantic alignment module (Q-Former), which utilizes a set of learnable query vectors to capture key information from visual features. Q-Former is trained through a vision-language matching task to align visual and textual features, enabling LLMs to generate textual descriptions based on visual input. 
To reduce reliance on paired image-text data, Jian et al. \cite{P} training their model on textual data to optimize prompts that guide the generation of language from visual inputs. Visual features are subsequently mapped to these prompts, achieving alignment between vision and language. 
To enhance the model’s understanding of fine-grained visual information, Lu et al. \cite{MQ} capture multi-level image features through three submodules: image tagging, object detection, and semantic segmentation. This enables the LLMs to process more granular visual information.  
Additionally, approaches \cite{Emotion-LLaMA} employ simple linear networks (e.g., MLP) to map non-textual features into the textual embedding space, facilitating modality alignment. 
These studies provide effective methods to extend text-only LLMs to process multimodal data.

\begin{figure*}[!h]
\centering
\includegraphics[width=0.95\linewidth]{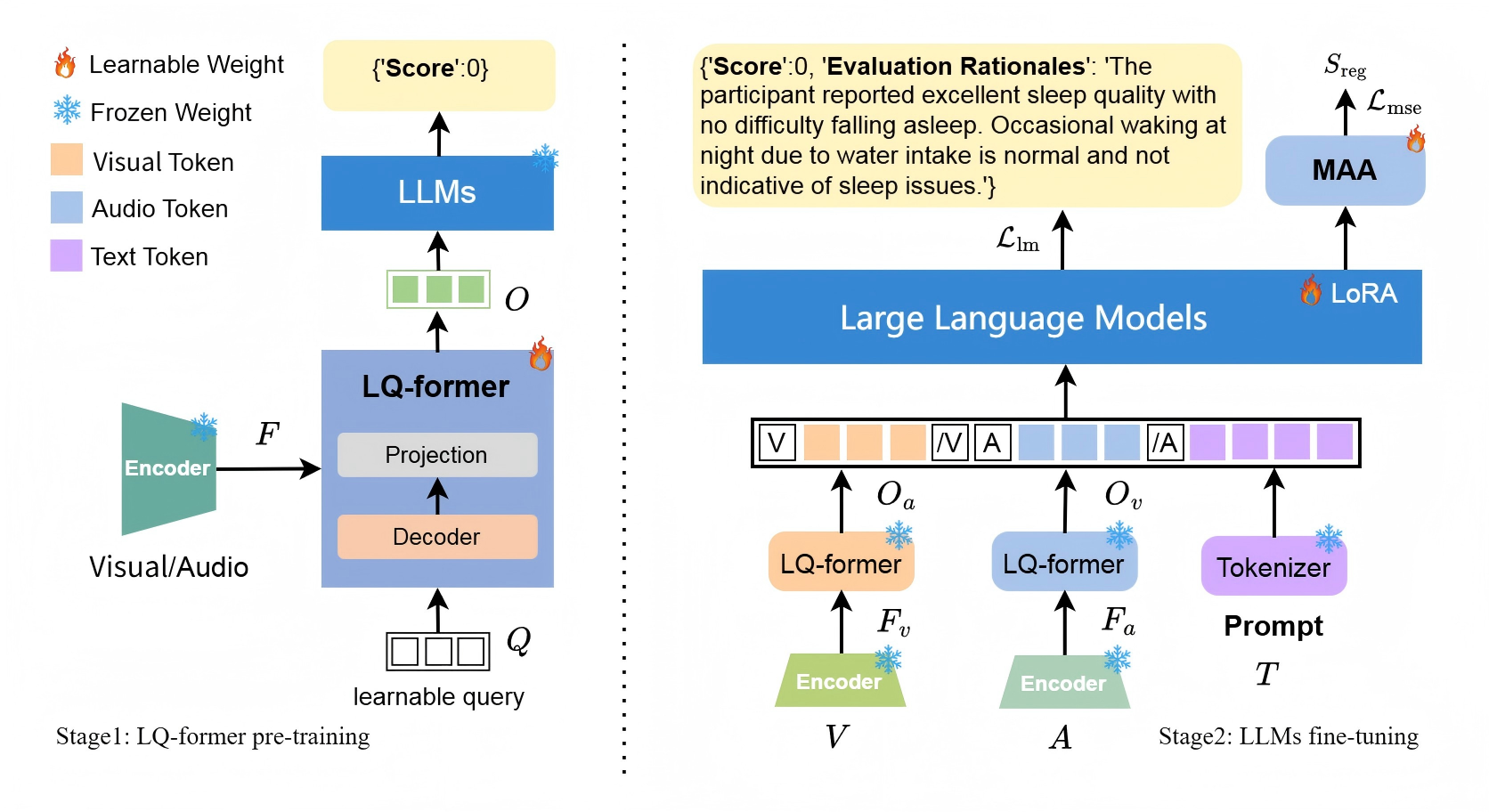}
\caption{The overall framework of the proposed MLlm-DR method. left shows the LQ-former pre-training part, which aims to extracts depression-related feature representations, comprehensible by LLMs, from visual and audio data. On the right is the LLMs fine-tuning part, where the learned feature representation is concatenated with text instruction embeddings as input to fine-tune the LLMs, which then output depression score and evaluation rationales. }
\label{fig_2}
\end{figure*}

\section{Methodology}

Our method takes a video clip \(C_i\) from the interview, corresponding to a specific psychological aspect \(i\) (e.g., sleep, appetite) of the participant, as input. 
Each \(C_i\) includes audio \(A\), visual \(V\), and transcribed text \(T\) data. The method outputs a score \(s_i\) reflecting the participant's depression score in aspect \(i\), along with the evaluation rationales. 
The final depression assessment score \(S\) is obtained by summing \(s_i\) across all psychological aspects, aligning closely with the clinical diagnostic process used by physicians.

As showing Figure ~\ref{fig_2}, the method consists of two stages: LQ-former pre-training and LLMs fine-tuning. In the LQ-former pre-training stage, visual or audio features \(F\) are first extracted using pre-extraction modules (Encoder). 
A learnable query vector \(Q\) is then used to learn a fixed-length features representation \(O\) from \(F\).  \(O\) is subsequently fed into the LLMs with frozen parameters to perform the score prediction task, thereby training LQ-former. 
In the LLMs fine-tuning stage, the parameters of the pre-trained LQ-former module are frozen. 
The audio features \(F_a \) and visual features \(F_v \) are then used to extract  \( O_a \) and \( O_v \), 
which are concatenated with text instruction embeddings as input to the LLMs.
Finally, the LLMs is fine-tuned using a joint optimization strategy that combines language model loss \(\mathcal{L}_{\text{lm}}\) and regression prediction loss \(\mathcal{L}_{\text{mse}}\), enabling the model to output both depression score and evaluation rationales.

\subsection{Data Processing}

To enhance the performance of 
small-scale LLMs
in depression assessment tasks and enable them to generate more logically consistent evaluation rationales, we utilize 
large-scale LLMs
\cite{gpt4,llama} to generate evaluation rationales based on dialogue content and construct training dataset. 
Specifically, we provide a text instruction requesting the 
large-scale LLMs
to produce an evaluation score within a range of 0-3 along with corresponding rationales based on the dialogue content. 
In this text instruction, we report the actual evaluation score of the patient, ensuring consistency between the rationales generated by the LLMs and the actual score, as shown in the Table \ref{label_1}.  
The training dataset is then used to fine-tune MLlm-DR, aiming to transfer the reasoning capabilities of the 
large-scale LLMs
to a smaller model for domain-specific tasks. 
In addition, we use HuBERT \cite{hubert} to extract audio features from raw speech data. For visual features, we utilize the deep representations provided in the dataset, which are extracted using OpenFace 2.0 \cite{OpenFace2.0} or ResNet 50 \cite{ResNet}.

\begin{table}[h]
\centering
\small
\setlength{\tabcolsep}{6pt}
\caption{Prompt Instructions}
\begin{tabular}{c p{8.6cm}} 
\toprule
\textbf{Role} & \multicolumn{1}{c}{\textbf{Content}} \\ \midrule 
System
& \textit{You are a psychiatrist, assessing the participant's mental health in certain aspects through a series of questions. A score of 0 means not at all, 1 means several days, 2 means more than half the days, and 3 means nearly every day.} 
\\ \midrule
User& \textit{Given the participant's self-rating score of \texttt{\{label\}}, please evaluate the participant's performance in \texttt{\{aspect\}} based on the dialogue content. The output format is as follows: Evaluation Reason: A concise and logical description. Each output must strictly follow this format, avoiding omissions or confusion.} 
\\ \toprule
\end{tabular}
\label{label_1}
\end{table}

\subsection{Lightweight Query Module}

The LQ-former is designed to help LLMs process and understand multimodal information. It consists of a Transformer \cite{attention} Decoder and a Projection Module based on a fully connected network. 
The decoder takes a set of learnable query vectors \( Q \) as input. In each decoder layer, the queries first interact with each other via self-attention, and then attend to the non-textual modality features (such as audio and visual embeddings) via cross-attention. In this way, the decoder encourages different queries to focus on different regions of the modality sequence and aggregate depression-related cues, producing fixed-length feature representations \( H \). 
\( H \) are then projected to the same dimensionality as the LLM's textual embeddings through a Projection layer,  producing the output \( O \), as shown in the Equation ~\ref{q_1}. 
In our task, we utilize two LQ-former modules to process audio and visual features separately. The resulting features are concatenated with the tokenized textual instruction embeddings and used as inputs to the LLMs. 
To indicate the positions of the inserted features, we employ two special tokens, \textless AudioHere\textgreater and \textless VideoHere\textgreater, as markers.

\begin{equation}
\label{q_1}
\begin{aligned}
H &= \text{softmax}\left(\frac{QK^{T}}{\sqrt{d_k}}\right) V ,
\\
O &= \text{Projection}(H),
\end{aligned}
\end{equation}
where \( Q \in{\mathbb{R}^{n\times{D}}}\) is the query vector, while \( K = V \in{\mathbb{R}^{m\times{D}}}\) represents the sequence of non-textual modality features, \( D \) is the embedding dimension, \( n\) denotes a fixed sequence length, and \( m \) represents the length of the non-textual feature sequence.

\subsection{Multi-Head Attention Aggregation Network}

To process the regression prediction task, we design a Multi-Head Attention Aggregation Network (MAA).
MAA takes the final hidden states \(X \in \mathbb{R}^{L \times D}\) of the LLMs as input. 
It first splits \(X\) into \(h\) subspaces corresponding to \(h\) attention heads, resulting in a tensor \(x' \in \mathbb{R}^{L \times h \times d}\), 
Each head then computes token-level attention weights \(\alpha_i \in \mathbb{R}^L\) using a shared fully connected layer. 
These weights are used to aggregate the token representations into a sentence-level embedding.
The embedding is passed through a fully connected layer to output the predicted scores \(S_{\text{reg}}\), which are used to calculate the regression loss, as shown in the Equation ~\ref{q_3}.

\begin{equation}
\label{q_3}
\begin{aligned}
\alpha_i &= \text{softmax}\left( \text{W}_i x_i'^\top \right), \\
S_{\text{reg}} &= \text{FC}\left( \sum_{i=1}^{h} \left( \alpha_i \cdot x_i' \right) \right),
\end{aligned}
\end{equation}
where \(L\) is the sequence length, \(h\) is the number of attention heads, and \(d = D / h\) is the dimension of each subspace, 
\(W_i \in \mathbb{R}^{d \times 1}\) is a trainable parameter for each head.

\subsection{Multi-Task Joint Fine-Tuning}

Existing LLMs primarily focus on text generation tasks, and their performance in score prediction is often limited when the provided prompt lacks sufficient example samples. 
To enhance the model's performance in scoring prediction tasks, we introduce a joint optimization strategy combining causal language modeling cross-entropy loss (\(\mathcal{L}_{\text{lm}}\)) and mean squared error loss (\(\mathcal{L}_{\text{mse}}\)), as shown in the Equation ~\ref{q_2}. 

\begin{equation}
\label{q_2}
\begin{aligned}
\mathcal{L}_{\text{lm}} &= - \frac{1}{T} \sum_{t=1}^{T} \log p_\theta(x_t \mid x_{<t}),\\
\mathcal{L}_{\text{joint}} &= \mathcal{L}_{\text{lm}} + \mathcal{L}_{\text{mse}},
\end{aligned}
\end{equation}
where $\mathcal{L}_{\text{mse}}$ is the mean squared error between the predicted score and the ground-truth score; $\mathcal{L}_{\text{lm}}$ is the autoregressive language modeling loss, i.e., the token at position $t$ ($x_t$) is predicted only from the preceding tokens $x_{<t}$. \(T\) is the sequence length, \(p_\theta(x_t \mid x_{<t})\) is the model's conditional probability of predicting \(x_t\) given its context.

By leveraging multi-task learning, we enable mutual reinforcement between tasks, improving the accuracy of regression prediction while generating explainable reasoning for evaluation scores.
Additionally, since LLMs may occasionally fail to follow the instructions and produce results in the required format, leading to missing prediction scores, the regression prediction score can serve as a supplementary result, enhancing the model's generalization ability.

\section{Experimental setup}

\subsection{Datasets}
We conduct experiments on two interview-based multimodal depression datasets (CMDC \cite{CMDC} and E-DAIC-WOZ \cite{DAIC}) to evaluate the proposed method. Both datasets provide raw speech and text data, as well as scale-based scoring results. A score $\geq$ 10 indicates that the participant is experiencing severe depression. Additionally, for patient privacy protection, both datasets only provide the extracted visual features. 
CMDC: A Chinese dataset containing 78 samples (26 depressed patients and 52 healthy individuals), with PHQ-9 used as the corresponding questionnaire. Some subjects did not complete the video recording, resulting in missing visual features.
E-DAIC-WOZ: An extended version of the DAIC-WOZ dataset, including 163 training samples, 56 validation samples, and 56 test samples, with PHQ-8 as the corresponding questionnaire.

Our model is designed to assess each aspect independently and then aggregate the results to predict the overall depression score. Therefore, we treat each aspect-level response as an individual training instance, resulting in a total of $78 \times 9 = 702$ training samples in CMDC and $163 \times 8 = 1304$ in E-DAIC-WOZ. For both datasets, we adopt subject-independent splits, ensuring that all samples from the same participant (e.g., different PHQ items) appear in only one split, thus avoiding information leakage.

In the CMDC, the 12 interview questions have clear correspondences with the 9 aspects of PHQ-9. For example, questions 4 and 6 correspond to the "loss of interest" aspect, questions 9 and 10 to "low mood", and question 8 to "self-harm or suicidal thoughts", among others. Therefore, our method selectively uses only the interview content that is clearly associated with each aspect to assess the participant's score on that aspect.
In contrast, in the E-DAIC-WOZ, the interview content primarily consists of a series of open-ended questions and does not have explicit correspondences with the items of PHQ-8. Thus, our method utilizes the complete interview content to assess the participant's depression score for each aspect.

\subsection{Implementation Details}

The method use LLaMA-3-8B \cite{llama} as the base model with fine-tuning and functional extensions, and GPT-4o  \cite{gpt4} to construct the training datasets. 
The speech features are deep representations of 768-dim, while the visual features are deep representations of 709-dim or 2048-dim. For missing sequence features, zero padding is applied. LQ-former uses 32 query vectors, each with a dimension of 768-dim, a hidden layer dimension of 1024-dim, and 4 network layers, with a dropout rate set to 0.3. The projection layer employs a two-layer fully connected network with a hidden layer dimension of 1024-dim and an output dimension of 4096-dim. MAA uses 8 fully connected networks to learn the sequence weights for each attention head, followed by a two-layer fully connected network for regression prediction. 
We fine-tune the query and value projection matrices ($W_q$ and $W_v$) using LoRA by setting $r = 16$, $\alpha = 32$ and dropout = 0.1. 
The learning rate is set to 0.00001, and the model is trained for 10 epochs. 
Text generation uses nucleus sampling with \texttt{do\_sample=True}, \texttt{temperature=0.8}, and \texttt{top\_k=10}.
All experiments are conducted using the PyTorch deep learning framework, and the training is performed on two A800 GPUs.

\subsection{Metrics}
To comprehensively evaluate the model's performance, we employed a variety of metrics. For binary classification tasks, Precision, Recall, and F1-Score were used to measure the accuracy of classification. For regression tasks, Concordance Correlation Coefficient (CCC), as shown in the Equation ~\ref{q_4}, Root Mean Squared Error (RMSE), and Mean Absolute Error (MAE) were utilized to assess the differences between predicted and actual values. These metrics provide a multidimensional evaluation, offering a comprehensive understanding of the model's performance.

\begin{equation}
\label{q_4}
\text{CCC} = \frac{2\rho\sigma_x\sigma_y}{\sigma_x^2 + \sigma_y^2 + (\mu_x - \mu_y)^2},
\end{equation}
where \( \rho \) represents the Pearson correlation coefficient, \( \sigma_x \) and \( \sigma_y \) denote the standard deviations of the ground truth and predicted values, respectively, and \( \mu_x \) and \( \mu_y \) are their corresponding means. CCC values range from -1 to 1, with 1 indicating perfect agreement.

\section{Experimental Results and Analysis}

\subsection{Baseline Methods}

In our experiments, we introduce the following baseline methods for comparison: 
{\bf{1) CubeMLP}}\cite{cubemlp}: Utilizes multilayer perceptrons (MLPs) to perform feature fusion across multiple dimensions, enabling comprehensive interactions between modalities.
{\bf{2) MulT}}\cite{mult}: A cross-attention-based multimodal fusion method designed to model interactions across different modalities. 
{\bf{3) MMFF}}\cite{MMFF}: Extracting multi-order factors from different modalities and their combinations, providing richer representational capacity for multimodal learning. 
{\bf{4) IIFDD}}\cite{iifdd}: Integrating intra-modality and inter-modality feature fusion frameworks for multimodal depression recognition. 
{\bf{5) HiQuE}}\cite{HiQuE}: Enhancing depression diagnosis accuracy by analyzing the importance weights of different questions within each modality.
{\bf{6) P+W+A}}\cite{Harnessing}: Using prompts to guide LLMs in extracting textual features from transcripts, which are then fused with other modality features (AU, pose, gaze) for multimodal fusion.
{{\bf 7) MMCL} \cite{MMCL}: captures enhanced and complementary features from both modality-common and modality-specific representations, which are then fused into a unified representation for emotion prediction.

\subsection{Comparative Experimental}

We compared our proposed method with the latest in inroaches on the CMDC and E-DAIC-WOZ interview-based depression datasets, as shown in Table ~\ref{label_2}. 
The results demonstrate that our method outperforms all existing schemes, achieving state-of-the-art performance. 
Specifically, on the CMDC dataset, our method achieved an exceptional 100\% in Precision, Recall, and F1 Score. 
On the E-DAIC-WOZ dataset, it also significantly outperformed other methods across all metrics, confirming the robustness and generalization ability of the proposed framework. 
This superior performance can be attributed to the LLM’s powerful contextual understanding capabilities, which enable it to capture fine-grained semantic and emotional cues from multi-turn dialogues, thereby allowing a more accurate analysis of the participant’s psychological state in specific aspects.

\begin{table*}[h!]
\centering
\caption{
Comparison of the performance of recent methods on the CMDC and E-DAIC-WOZ datasets, Bold numbers indicate the best performance.}
\begin{tabular}{cccccccc}
\toprule
\textbf{Dataset} & \textbf{Model}& \textbf{CCC} $\uparrow$& \textbf{Pre} $\uparrow$& \textbf{Rec} $\uparrow$& \textbf{F1} $\uparrow$& \textbf{RMSE} $\downarrow$& \textbf{MAE} $\downarrow$
\\ \midrule
\multirow{7}{*}{CMDC} 
& MulT(2019) & \textbackslash & 0.72 & \textbackslash  & \textbackslash & 4.59 & 3.66 \\
& CubeMLP(2022) & \textbackslash & 0.38 & 0.8 & 0.51 & \textbackslash & \textbackslash \\ 
& MMFF(2022) & \textbackslash & 0.83 & \textbackslash  & \textbackslash & 4.29 & 3.19 \\
& IIFDD(2023) & \textbackslash & 0.95 & 0.93 & 0.93 & \textbackslash & \textbackslash \\
& MMCL(2025) & \textbackslash & 1.00 & 0.93 & 0.96 & 3.43 & \textbf{2.31} \\
& Ours & \textbf{0.91}& \textbf{1.00}& \textbf{1.00}& \textbf{1.00} & \textbf{3.10} & 2.61 
\\ \midrule
\multirow{7}{*}{\makecell[c]{E-DAIC-\\WOZ}} 
& MulT(2019) & \textbackslash & 0.64& 0.65& 0.64& \textbackslash&\textbackslash
\\
& CubeMLP(2022) & 0.58& \textbackslash& \textbackslash& \textbackslash& \textbackslash&4.37
\\
& MMFF(2022) & 0.67& \textbackslash& \textbackslash  & \textbackslash & 4.91&3.98\\
& HiQuE(2024)& \textbackslash & 0.71 & 0.70& 0.70& \textbackslash &\textbackslash \\
& P+W+A(2024)& \textbackslash& \textbackslash& \textbackslash& \textbackslash& 4.66&3.86\\
& MMCL(2025) & \textbackslash & 0.74 & 0.73 & 0.72 & 4.76 & 3.85 \\
& Ours & \textbf{0.72}& \textbf{0.77}& \textbf{0.81}& \textbf{0.79}& \textbf{4.59}&\textbf{3.41}\\
\toprule
\end{tabular}
\label{label_2}
\end{table*}

Compared with non-LLMs-based methods such as MulT~\cite{mult}, CubeMLP~\cite{cubemlp}, MMFF~\cite{MMFF}, IIFDD~\cite{iifdd}, HiQuE~\cite{HiQuE}, and MMCL~\cite{MMCL}, our model exhibits substantial advantages. 
These methods typically rely on small language models such as BERT~\cite{BERT} for textual feature extraction, which limits their contextual awareness in multi-turn dialogues and consequently constrains their overall performance.
Compared with LLMs-based approaches such as P+W+A~\cite{Harnessing}, our method still achieves superior overall results. P+W+A first uses an LLMs to extract textual representations and then performs fusion with acoustic and visual embeddings. In contrast, our method retains its advantage by integrating multimodal information and LLMs within a unified framework, reducing potential information loss in staged processing pipelines.

\subsection{Effectiveness of the Training Dataset}

To evaluate the effectiveness of the constructed training dataset, we conduct comparative experiments on two baseline methods, as shown in Table~\ref{label_3}.
{\bf{1) LLaMA-3-8B}}\cite{llama}: Directly using the LLaMA-3-8B model to analyze dialogue content and generate evaluation results. 
{\bf{2) LoRA}\cite{lora}}: Fine-tuning the LLaMA-3-8B with the LoRA method using the training dataset.

\begin{table}[h!]
\centering
\caption{Comparison of the performance of different baselines on the CMDC and E-DAIC-WOZ datasets (i.e., \{T\} indicates that only the text modality was used).}
\begin{tabular}{cccccccc}
\toprule
\textbf{Dataset} & \textbf{Model}  &\textbf{CCC} $\uparrow$& \textbf{Pre} $\uparrow$& \textbf{Rec} $\uparrow$& \textbf{F1} $\uparrow$& \textbf{RMSE} $\downarrow$& \textbf{MAE} $\downarrow$ 
\\
\midrule
\multirow{3}{*}{\makecell[c]{CMDC}}
& LLaMA-3-8B \{T\}& 0.37 & 0.35 & 1.00 & 0.52 & 10.19&9.36\\
& LoRA \{T\}& 0.80 & 1.00 & 0.83 & 0.91 & 4.61 &4.08 \\ 
& Ours & \textbf{0.91}& \textbf{1.00}& \textbf{1.00}& \textbf{1.00} & \textbf{3.10} & \textbf{2.61} \\
\midrule
\multirow{3}{*}{\makecell[c]{E-DAIC-\\WOZ}} 
& LLaMA-3-8B \{T\}& 0.54& 0.56& \textbf{0.85}& 0.68& 5.55&4.23\\
& LoRA \{T\} & 0.69& 0.74& 0.80& 0.77& 5.04& 4.04\\
& Ours& 
\textbf{0.72}& \textbf{0.77}& 0.81& \textbf{0.79}& \textbf{4.59}& \textbf{3.41} \\ 
\toprule
\end{tabular}
\label{label_3}
\end{table}

Directly using LLaMA-3-8B for prediction does not perform well, primarily due to three factors:
1) the limited logical reasoning capabilities of 
small-scale LLMs
, which make it challenging to provide accurate evaluations based on dialogue content; 
2) the model often fails to strictly follow instruction-defined scoring formats, leading to unusable outputs; 
3) the inherent limitations of language models in regression prediction tasks, stemming from a lack of domain-specific knowledge and sufficient labeled sample guidance. 
However, these issues were significantly mitigated through LoRA-based fine-tuning on training datasets, leading to substantial performance improvements.

\subsection{Ablation Study}

To verify the effectiveness of the LQ-former module and the multi-task learning strategy, we conduct extensive ablation experiments, as shown in Table ~\ref{label_4}. 
The experimental settings include: 
{\bf{1) w/o Joint}}: Uses the LQ-former module to integrate audio and visual information and fine-tunes the LLMs, without multitask learning.
{\bf{2) w/o LQ}}: Excludes the LQ-former module, not integrating audio and visual information, but fine-tunes with joint optimization.
{\bf{3) w/o LQ-A}}: Excludes the audio LQ-former module but retains the visual LQ-former module, fine-tuning with joint optimization.
{\bf{4) w/o LQ-V}}: Excludes the visual LQ-former module but retains the audio LQ-former module, fine-tuning with joint optimization.

Experimental results demonstrate that the LQ-former module and the multi-task learning strategy can significantly enhance model performance, and removing either one would result in a performance decline.
The LQ-former is capable of extracting depression-related cues from acoustic and visual data, thereby providing the LLMs with more informative cross-modal representations. 
Meanwhile, the joint learning strategy introduces a regression prediction task that enables the LLMs to generate scores more closely aligned with the true values, even when prompt examples are insufficient.
Additionally, removing either audio or visual data in the LQ-former module also leads to a slight decline in model performance.

\begin{table}[h!]
\centering
\caption{Result of ablation study on the CMDC and E-DAIC-WOZ datasets.}
\begin{tabular}{cccccccc}
\toprule
\textbf{Dataset} & \textbf{Model}  &\textbf{CCC} $\uparrow$& \textbf{Pre} $\uparrow$& \textbf{Rec} $\uparrow$& \textbf{F1} $\uparrow$& \textbf{RMSE} $\downarrow$& \textbf{MAE} $\downarrow$ 
\\
\midrule
\multirow{5}{*}{\makecell[c]{CMDC}}& w/o Joint&  0.89& 1.00& 1.00& 1.00& 3.58& 3.08\\
& w/o LQ& 0.87& 1.00& 1.00& 1.00& 4.18& 3.61\\
& w/o LQ-A& 0.87& 1.00& 1.00& 1.00& 3.99&3.55\\
& w/o LQ-V& 0.90& 1.00& 1.00& 1.00& 3.52&2.90\\
& Ours& \textbf{0.91}& \textbf{1.00}& \textbf{1.00}& \textbf{1.00}& \textbf{3.10}& \textbf{2.61}\\ 
\midrule
\multirow{5}{*}{\makecell[c]{E-DAIC-\\WOZ}} & w/o Joint
& 0.67& 0.75& 0.71& 0.73& 4.74& 3.58\\
& w/o LQ& 0.65& 0.67& 0.76& 0.71& 4.86& 3.78\\
& w/o LQ-A& 0.66& 0.70& 0.72& 0.73& 4.80&3.71\\
& w/o LQ-V& 0.70& 0.75& 0.79& 0.77& \textbf{4.55}&3.56\\
& Ours& 
\textbf{0.72}& \textbf{0.77}& \textbf{0.81}& \textbf{0.79}& 4.59& \textbf{3.41}\\ 
\toprule
\end{tabular}
\label{label_4}
\end{table}

\subsection{Analysis of LQ-former}

To validate the effectiveness of the features extracted by the LQ-former module from visual and speech data, we report the pre-training results of the LQ-former module,  as shown in Table ~\ref{label_6}. 
The feature representations extracted by LQ-former are fed into a a froze LLMs to generate depression scores, in the format of "Score: 0". The quality of the generated results reflects the LQ-former module's performance in two aspects: first, its ability to extract depression-related features from speech and visual data; second, whether these features can be effectively understood by the LLMs. We did not have the model output the corresponding evaluation rationales, as generating fine-grained explanations solely based on speech and visual information is challenging. The main reason is that speech and visual data lack large-scale, fine-grained emotional labels, which prevents the LLMs from generating detailed and accurate explanations as it would with text data.

We evaluate three configurations: {\bf1) {LQ-A}}: Extract depression-related features from speech data using the LQ-former. {\bf2) {LQ-V}}: Extract depression-related features from visual data using the LQ-former. {\bf3) {LQ}}: Extract depression-related features from both speech and visual data using the LQ-former.
The experimental results show that the LQ-former effectively extracts depression-related feature representations from audio and visual data that can be understood by LLMs, with this phenomenon being particularly pronounced in the CMDC dataset.
The reason for this difference lies in the task complexity: in the CMDC dataset, the LQ-former extracts depression-related features from specific interview segments (averaging 1 minute), making the task simpler and yielding better performance. In contrast, in the E-DAIC-WOZ dataset, the LQ-former must extract depression-related features from the entire interview content (averaging 20 minutes), which increases the task difficulty and results in poorer performance.

\begin{table}[h!]
\centering
\caption{Result of LQ-former study on the CMDC and E-DAIC-WOZ datasets.}
\begin{tabular}{cccccccc}
\toprule
\textbf{Dataset} & \textbf{Model} &\textbf{CCC} $\uparrow$& \textbf{Pre} $\uparrow$& \textbf{Rec} $\uparrow$& \textbf{F1} $\uparrow$& \textbf{RMSE} $\downarrow$& \textbf{MAE} $\downarrow$ 
\\ \midrule
\multirow{3}{*}{\makecell[c]{CMDC}}& LQ-A
& 0.81& 0.94& 1.00& 0.97& 4.01& 3.26\\
& LQ-V
& 0.29& 0.60&  0.50& 0.55& 8.13&  6.89\\
& LQ& 0.85& 1.00& 1.00& 1.00& 3.79& 3.06\\ 
\midrule
\multirow{3}{*}{\makecell[c]{E-DAIC-\\WOZ}} 
& LQ-A&  0.09&  0.45& 0.42& 0.43& 7.81& 6.48\\
& LQ-V&  0.07& 0.50& 0.42&  0.45& 7.88&6.18\\
& LQ& 0.22& 0.52& 0.50& 0.51& 7.24& 6.38\\ \toprule
\end{tabular}
\label{label_6}
\end{table}

\subsection{Analysis of Evaluation Rationales}

To further evaluate the interpretability and reliability of the model's outputs, we conducted a human evaluation involving clinical experts. 
We randomly selected 100 samples from the test set and invited three assessors with mental-health–related professional backgrounds who had received formal training from physicians to perform independent reviews. 
The experts evaluated the corresponding model-generated rationales.
Each rationale was rated on a 3-point scale: 3 — \textit{Fully agree} (the expert would have made the same assessment), 2 — \textit{Reasonable} (different perspective but similar conclusion), and 1 — \textit{Disagree} (the rationale was not acceptable). The final per-sample rationale label was determined by majority vote; in the event of a tie, the median rating was used.
We report four evaluation metrics in Table ~\ref{label_7}:
I) The proportion of model outputs not conforming to the required instruction format;
II) agreement rate between model predictions and expert ratings;
III) Result of expert ratings (RMSE/MAE);
IV) Proportional distribution of expert ratings on the quality of model-generated evaluation rationales (3/2/1).

\begin{table}[h!]
\centering
\caption{Results of Human Expert Evaluations}
\begin{tabular}{ccccccc}
\toprule
\multirow{3}{*}{Methods}& \multicolumn{3}{c}{CMDC} & \multicolumn{3}{c}{E-DAIC-WOZ} \\
\cmidrule(lr){2-4} \cmidrule(lr){5-7}
& \makecell[c]{LLaMA-3-8B}& LoRA& Ours& \makecell[c]{LLaMA-3-8B}& LoRA& Ours\\
\midrule
I&9.26\%& 0.61\%& 1.23\%& 18.52\%& 4.32\%&3.12\%
\\
II& \multicolumn{3}{c}{87\%}& \multicolumn{3}{c}{73\%}\\
III& \multicolumn{3}{c}{3.25/2.21}& \multicolumn{3}{c}{5.01/3.94}\\
IV& \multicolumn{3}{c}{77\%/14\%/9\%}& \multicolumn{3}{c}{64\%/19\%/17\%}\\
\toprule
\end{tabular}
\label{label_7}
\end{table}

From the results, we draw the following observations:
{\bf{1)}} LLaMA-3-8B frequently produced improperly formatted outputs, impairing scoring accuracy, while LoRA fine-tuning effectively mitigated this issue and enhanced model performance.
{\bf{2)}} Our method achieved 87\% and 73\% consistency with expert scores on CMDC and E-DAIC-WOZ respectively, demonstrating strong practical value. 
{\bf{3)}} Expert ratings also showed limitations, even falling slightly behind our method on E-DAIC-WOZ, reflecting the inherent difficulty of depression assessment and the subtlety of depressive cues that often require additional context to interpret. 
{\bf (4)} On CMDC and E-DAIC-WOZ, 77\% and 64\% of the model’s rationales were rated “fully agree” by clinical experts, further corroborating the interpretability and clinical concordance of the proposed approach.

\subsection{Case Study}

To gain deeper insights into the reasoning behaviors of different models, we conducted a qualitative case study on two representative samples from the CMDC and E-DAIC-WOZ datasets, as shown in Figure~\ref{fig_3}. 
We focus on three aspects: the consistency between predicted scores and evaluation rationale, the relevance of reasoning to original dialogue, and the logical soundness of the reasoning. 

\begin{figure*}[!h]
\centering
\includegraphics[width=1\linewidth]{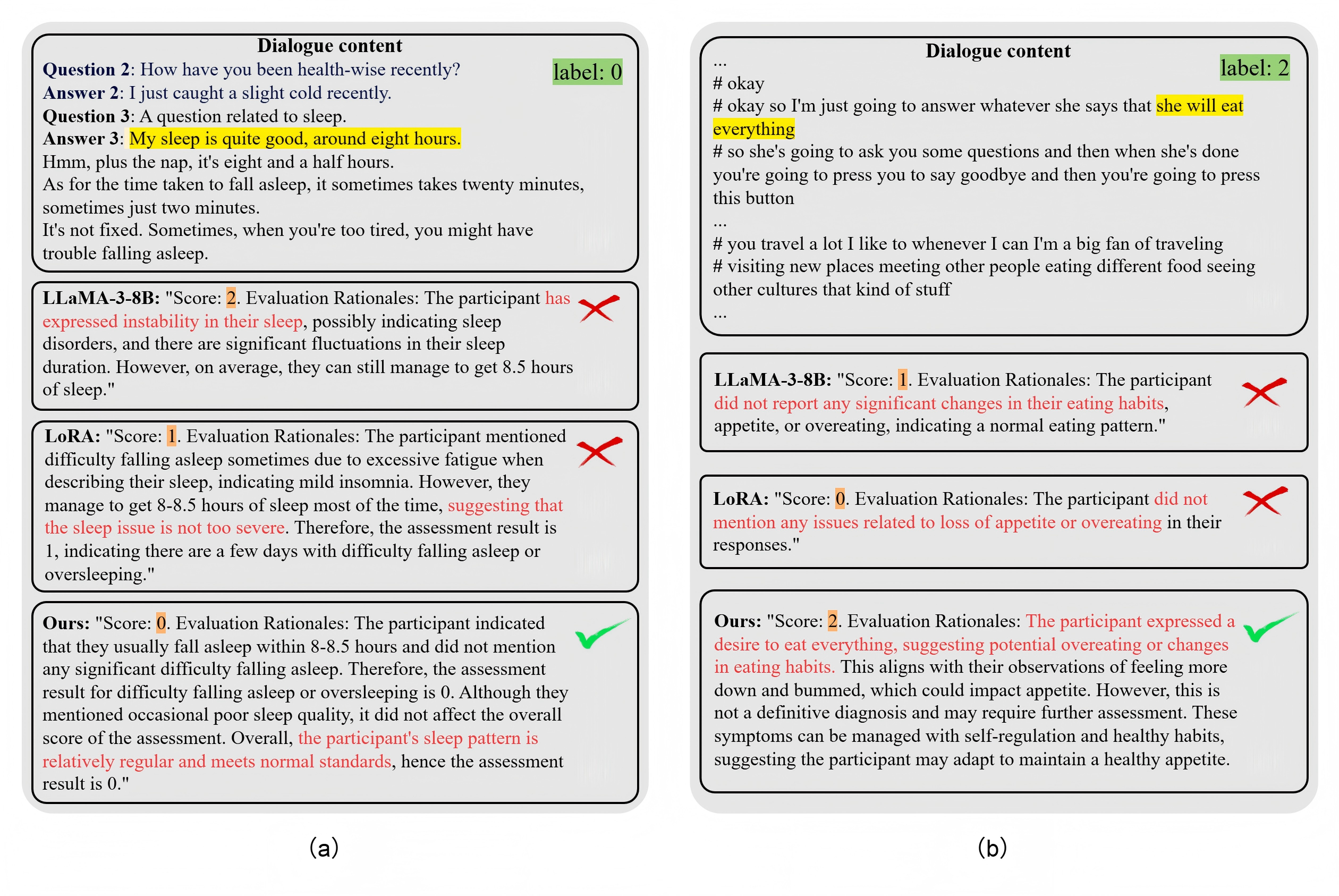}
\caption{
Case analysis of explainable depression recognition on the (a) CMDC and (b) E-DAIC-WOZ datasets. We present the inference results from three different models. ``Dialogue content" refers to excerpts from the interview process, ``label" represents the participant's true score, the highlighted orange section represents the model's predicted score, the highlighted yellow sections indicate the key parts of the dialogue content, and the red text indicates the key explanations in the evaluation rationale that are related to the corresponding scores. }
\label{fig_3}
\end{figure*}

The results reveal a progressive improvement across methods:
1) {\bf{LLaMA-3-8B}} method showed clear deficiencies in score prediction. For instance, although the model recognized that the subject had 8 hours of sleep, it still gave a score of 2 (poor), which did not align with the actual situation. 2) {\bf{LoRA}} method {fine-tuned model} showed improvements, providing a "mild" reasoning result, but the gap between the predicted scores and actual values remained significant. This limitation is primarily due to the lack of real labels during training, leading to deviations between predicted and actual scores. 3) {\bf{Our}} method demonstrated a clear advantage in capturing subtle cues in the dialogue that are difficult to detect, enabling accurate predictions. 
This advantage is attributed to the integration of multimodal and multi-task learning modules, which provide the model with latent information, thereby enhancing the accuracy of score predictions.

\subsection{Efficiency Analysis}

We further analyze the efficiency of our method. Specifically, we adopt a two-stage training strategy. In Stage 1, the LLMs is frozen and only the LQ-former is trained. The LQ-former consists of four Transformer decoder layers (hidden size = 1024) with 32 learnable queries and a two-layer projection module. The additional computational overhead in this stage primarily comes from multimodal feature encoding. In Stage 2, we introduce LoRA adaptation to the ($W_q$ and $W_v$) matrices of LLaMA-3-8B with ($r = 16$) and ($\alpha = 32$). The number of trainable parameters is on the order of several million, which is three orders of magnitude smaller than full-parameter fine-tuning ($\approx$8B parameters). Consequently, the overall training cost is only slightly higher than that of standard LoRA-based baselines, and during inference our method adds merely a single forward pass of the LQ-former, whose latency is negligible compared to the LLMs. A detailed comparison of trainable parameters, GPU memory, and training time is provided in Table~\ref{label_12}.

Despite this minimal extra cost, our method yields substantial performance improvements over the LoRA baseline, as shown in Table~\ref{label_3}. On E-DAIC-WOZ, CCC increases from 0.69 to 0.72, F1 from 0.77 to 0.79, and MAE decreases from 4.04 to 3.41; on CMDC, RMSE drops from 4.61 to 3.10, MAE from 4.08 to 2.61, and all three classification metrics reach 1.00. These results demonstrate that the proposed two-stage strategy offers a highly favorable performance–efficiency trade-off, delivering stable and substantial gains with only marginal additional training overhead.

\begin{table}[!h]
\centering
\small
\caption{Comparison of training cost between different models on the CMDC dataset.}
\label{label_12}
\begin{tabular}{clccc}
\toprule
\textbf{Dataset} & \textbf{Model} & \textbf{Trainable Params} & \textbf{GPU Memory} & \textbf{Training Time} \\
\midrule
\multirow{3}{*}{CMDC}
& MulT            & 20M & 16G & 1.5h \\
& LLaMA-3--8B     &  8M & 80G & 2.5h \\
& Ours            & 25M & 80G & 3.0h \\
\bottomrule
\end{tabular}
\end{table}

\subsection{Model Scalability and Efficiency}

To analyze the impact of model size, we instantiated MLlm-DR with LLaMA-3 models of different parameter scales (1B, 3B, and 8B) while keeping all training settings and the LQ-former unchanged. As shown in Table~\ref{label_8}, performance declines consistently as the model size decreases. 
The 3B model exhibits only a mild degradation relative to the 8B baseline, whereas the 1B model shows substantial drops across all metrics, with CCC and F1 decreasing by 25.3\% and 61.1\%, respectively. These findings indicate that an 8B-scale LLMs provides a favorable balance between reasoning capability and computational overhead for explainable depression recognition, while excessively small models are insufficient to capture the nuanced semantic and emotional cues present in clinical interviews.

\begin{table}[h!]
\centering
\caption{Performance of MLlm-DR with different scales LLM on the E-DAIC-WOZ dataset.}
\begin{tabular}{cccccccc}
\toprule
\textbf{Dataset} & \textbf{Model} &\textbf{CCC} $\uparrow$& \textbf{Pre} $\uparrow$& \textbf{Rec} $\uparrow$& \textbf{F1} $\uparrow$& \textbf{RMSE} $\downarrow$& \textbf{MAE} $\downarrow$ 
\\ \midrule
\multirow{3}{*}{\makecell[c]{E-DAIC-\\WOZ}} 
& LLaMA-3-1B & 0.28 & 0.52 & 0.71 & 0.59 & 6.05 & 4.94 \\
& LLaMA-3-3B & 0.65 & 0.76 & 0.54 & 0.63 & 4.70 & 3.59 \\
& LLaMA-3-8B & 0.72 & 0.77 & 0.81 & 0.79 & 4.59 & 3.41 \\
\toprule
\end{tabular}
\label{label_8}
\end{table}

\subsection{Cross-Corpus Generalization Study}

To further verify the generalization capability of MLlm-DR, we conducted a cross-corpus experiment training on the Chinese CMDC dataset and evaluating on the English E-DAIC-WOZ dataset. This setting challenges the model not only on domain shift but also on linguistic diversity. Specifically, to bridge the linguistic gap, the text modality was translated to align with the target language during the evaluation. As shown in Table ~\ref{label_9}, our model demonstrates robust performance in this challenging setting, outperforming the zero-shot baseline of LLaMA-3-8B. This confirms that our method possesses strong generalization capabilities across different languages and data distributions, even when constrained by feature heterogeneity.

\begin{table}[h!]
\centering
\caption{Cross-corpus evaluation: training on CMDC and testing on E-DAIC-WOZ.}
\begin{tabular}{ccccccc}
\toprule
\textbf{Model} &\textbf{CCC} $\uparrow$& \textbf{Pre} $\uparrow$& \textbf{Rec} $\uparrow$& \textbf{F1} $\uparrow$& \textbf{RMSE} $\downarrow$& \textbf{MAE} $\downarrow$ 
\\ \midrule
zero-shot & 0.54 & 0.56 & 0.85 & 0.68 & 5.55 & 4.23 \\
cross-corpus & 0.69 & 0.65 & 0.71 & 0.68 & 5.04 & 4.04 \\
\toprule
\end{tabular}
\label{label_9}
\end{table}

\subsection{Effect of Different Speech Features}

To investigate how different acoustic representations affect the multimodal reasoning ability of MLlm-DR, we replaced the original HuBERT features with two alternative extractors, Mel-spectrograms\cite{mcfee2015librosa} and WavLM\cite{chen2022wavlm}, and conducted controlled experiments on the CMDC dataset. As shown in Table~\ref{label_10}, Mel-spectrogram features lead to a marked degradation across all metrics, suggesting that low-level spectral energy information alone is insufficient to capture depression-related speech patterns. WavLM outperforms Mel and yields moderate gains in CCC and RMSE, indicating that its self-supervised acoustic representations are more effective for modeling speaker state; however, its overall performance still falls short of HuBERT. HuBERT, trained with a “cluster + masked prediction’’ objective, preserves higher-level speech representations that are tightly coupled with semantic content while still encoding paralinguistic cues. Therefore, HuBERT remains the most suitable backbone for speech feature extraction in our clinical interview setting.

\begin{table}[h!]
\centering
\caption{Performance of MLlm-DR with different acoustic feature on the CMDC dataset.}
\begin{tabular}{clcccccc}
\toprule
\textbf{Dataset}  &Model&\textbf{CCC} $\uparrow$& \textbf{Pre} $\uparrow$& \textbf{Rec} $\uparrow$& \textbf{F1} $\uparrow$& \textbf{RMSE} $\downarrow$& \textbf{MAE} $\downarrow$ 
\\ \midrule
\multirow{3}{*}{\makecell[c]{CMDC}} 
 &Mel& 0.40 & 0.55 & 0.83 & 0.67 & 7.78 & 6.28 \\
 &WavLM& 0.59 & 0.71 & 0.83 & 0.79 & 6.12 & 4.83 \\
 &HuBERT& 0.81 & 0.94 & 1.00 & 0.97 & 4.01 & 3.26 \\
\toprule
\end{tabular}
\label{label_10}
\end{table}

\subsection{Stability and Robustness Analysis}

To further evaluate the robustness and stability of MLlm-DR, we conducted five independent runs on both the CMDC and E-DAIC-WOZ datasets using the same LLaMA-3.2-8B backbone and identical training configurations. As summarized in Table~\ref{label_11}, the model maintains consistently high performance across runs, with only minor fluctuations in CCC, F1, RMSE, and MAE. On CMDC, the average CCC reaches 0.926 and F1 remains at 0.964, demonstrating strong stability on interview data. On E-DAIC-WOZ, despite the dataset’s open-ended conversational nature, the model still achieves an average CCC of 0.646 and F1 of 0.688, confirming its reliable generalization behavior. The results confirm that the performance gains of MLlm-DR are stable and statistically meaningful, indicating that they stem from the method itself rather than stochastic fluctuations.

\begin{table}[h!]
\centering
\caption{Results of five independent runs of MLlm-DR on the CMDC and E-DAIC-WOZ dataset.}
\begin{tabular}{cccccccc}
\toprule
\textbf{Dataset} & \textbf{Run} & \textbf{CCC} $\uparrow$ & \textbf{Pre} $\uparrow$ & \textbf{Rec} $\uparrow$ & \textbf{F1} $\uparrow$ & \textbf{RMSE} $\downarrow$ & \textbf{MAE} $\downarrow$ \\
\midrule
\multirow{6}{*}{CMDC} & 1 & 0.93 & 1.00 & 0.83 & 0.91 & 3.59 & 2.83 \\
& 2 & 0.93 & 1.00 & 0.83 & 0.91 & 3.63 & 2.94 \\
& 3 & 0.92 & 1.00 & 1.00 & 1.00 & 3.77 & 3.22 \\
& 4 & 0.94 & 1.00 & 1.00 & 1.00 & 3.55 & 2.89 \\
& 5 & 0.91 & 1.00 & 1.00 & 1.00 & 3.10 & 2.61 \\
& average & 0.93 & 1.00 & 0.93 & 0.96 & 3.52 & 2.89 \\
\midrule
\multirow{6}{*}{\makecell[c]{E-DAIC-\\WOZ}} & 1 & 0.64 & 0.80 & 0.67 & 0.73 & 4.77 & 3.58 \\
& 2 & 0.60 & 0.71 & 0.63 & 0.67 & 5.11 & 3.69 \\
& 3 & 0.67 & 0.72 & 0.54 & 0.62 & 4.66 & 3.46 \\
& 4 & 0.60 & 0.76 & 0.54 & 0.63 & 5.02 & 3.70 \\
& 5 & 0.72 & 0.77 & 0.81 & 0.79 & 4.59 & 3.41 \\
& average & 0.65 & 0.75 & 0.64 & 0.69 & 4.83 & 3.57 \\
\bottomrule
\end{tabular}
\label{label_11}
\end{table}

\subsection{Analysis of Data Collection Methods}

Our method shows significant performance differences between the CMDC and E-DAIC-WOZ datasets. 
In the CMDC dataset, the questions in the dialogue content are specifically designed based on the PHQ-9 scale, with each question having a clear correspondence to the participant's specific psychological state. We leverage this correspondence to select content related to the scale's questions and assess the participant's corresponding psychological state. This approach aligns closely with the depression diagnosis process used by clinicians based on interviews and has achieved outstanding results.
In contrast, the interview content in the E-DAIC-WOZ dataset is open-ended and lacks a fixed format, significantly increasing the complexity of evaluation. This difference limits the performance of existing models on the E-DAIC-WOZ dataset and highlights the critical role of data collection methods in automated depression recognition. This observation also provides valuable insights into optimizing data collection strategies in this field.

\section{Limitations}

Our model currently generates free-form textual descriptions to explain the patient's mental state, which may raise potential ethical and privacy concerns~\cite{zou2024weakly,liu2025prompting,wang2024q,hu2025multi}. 
In future work, we plan to incorporate domain knowledge and contextual constraints to guide the explanation generation process, thereby improving the consistency and safety of explanations. 
For example, adopting template-based or semi-structured generation mechanisms can enhance the controllability and auditability of model explanations while maintaining readability. 
In addition, in multimodal fusion tasks, low-quality or distorted modalities may reduce the model's ability to extract reliable features, leading to biased or unstable reasoning results~\cite{sun2022graphiqa,yu2025text}. 
In future studies, we will explore integrating perceptual quality assessment~\cite{yu2025dvlta} and cross-modal enhancement mechanisms~\cite{cheng20252} to improve the robustness and reasoning stability of the model.

\section{Conclusion}

In this paper, we propose a novel multimodal large language model (MLlm-DR). The model consists
of a small-scale LLMs and a lightweight query module (LQ-former), which are designed to generate
explainable evaluation rationales and integrate multimodal data, respectively, enabling explainable
and comprehensive depression diagnosis. This approach is closely aligned with clinical needs and
holds significant practical application value. Our approach achieves state-of-the-art results on two
interview-based benchmark datasets (CMDC and E-DAIC-WOZ), demonstrating its effectiveness and superiority. 
Beyond performance gains, this work advances the integration of LLMs and multimodal affective computing. The unification of language-driven reasoning and multimodal perception in MLlm-DR provides an interpretable and extensible basis for trustworthy clinical decision support.

\section{Acknowledgments}

This work is supported by National Science and Technology Innovation 2030 Major Project under Grant No. 2022ZD0209103. 

\bibliographystyle{plain}
\bibliography{sample-base}

@book{DSM-5,
  title={Diagnostic and statistical manual of mental disorders: DSM-5},
  author={American Psychiatric Association, DSMTF and American Psychiatric Association, DS and others},
  volume={5},
  number={5},
  year={2013},
  publisher={American psychiatric association Washington, DC}
}

@misc{ICD-11,
  title={International classification of diseases for mortality and morbidity statistics (11th Revision)},
  author={World Health Organization and others},
  year={2018}
}

@book{CCMD-3,
  title={Chinese Classification of Mental Disorders, 3rd Edition (CCMD-3)},
  author={Chinese Society of Psychiatry},
  publisher={Shandong Science and Technology Press},
  year={2001},
  address={Jinan, China}
}

@inproceedings{Q,
  title={Blip-2: Bootstrapping language-image pre-training with frozen image encoders and large language models},
  author={Li, Junnan and Li, Dongxu and Savarese, Silvio and Hoi, Steven},
  booktitle={International conference on machine learning},
  pages={19730--19742},
  year={2023},
  organization={PMLR}
}

@article{P,
  title={Bootstrapping vision-language learning with decoupled language pre-training},
  author={Jian, Yiren and Gao, Chongyang and Vosoughi, Soroush},
  journal={Advances in Neural Information Processing Systems},
  volume={36},
  pages={57--72},
  year={2024}
}

@article{MQ,
  title={Lyrics: Boosting fine-grained language-vision alignment and comprehension via semantic-aware visual objects},
  author={Lu, Junyu and Gan, Ruyi and Zhang, Dixiang and Wu, Xiaojun and Wu, Ziwei and Sun, Renliang and Zhang, Jiaxing and Zhang, Pingjian and Song, Yan},
  journal={arXiv preprint arXiv:2312.05278},
  year={2023}
}

@article{cogvlm,
  title={Cogvlm: Visual expert for pretrained language models},
  author={Wang, Weihan and Lv, Qingsong and Yu, Wenmeng and Hong, Wenyi and Qi, Ji and Wang, Yan and Ji, Junhui and Yang, Zhuoyi and Zhao, Lei and XiXuan, Song and others},
  journal={Advances in Neural Information Processing Systems},
  volume={37},
  pages={121475--121499},
  year={2024}
}

@article{fado,
  title={Fado: Feedback-aware double controlling network for emotional support conversation},
  author={Peng, Wei and Qin, Ziyuan and Hu, Yue and Xie, Yuqiang and Li, Yunpeng},
  journal={Knowledge-Based Systems},
  volume={264},
  pages={110340},
  year={2023},
  publisher={Elsevier}
}

@article{flamingo,
  title={Flamingo: a visual language model for few-shot learning},
  author={Alayrac, Jean-Baptiste and Donahue, Jeff and Luc, Pauline and Miech, Antoine and Barr, Iain and Hasson, Yana and Lenc, Karel and Mensch, Arthur and Millican, Katherine and Reynolds, Malcolm and others},
  journal={Advances in neural information processing systems},
  volume={35},
  pages={23716--23736},
  year={2022}
}

@inproceedings{sub-attentional,
  title={Multi-modal depression estimation based on sub-attentional fusion},
  author={Wei, Ping-Cheng and Peng, Kunyu and Roitberg, Alina and Yang, Kailun and Zhang, Jiaming and Stiefelhagen, Rainer},
  booktitle={European Conference on Computer Vision},
  pages={623--639},
  year={2022},
  organization={Springer}
}

@inproceedings{MMFF,
  title={Depression diagnosis and analysis via multimodal multi-order factor fusion},
  author={Yuan, Chengbo and Liu, Xuxu and Xu, Qianhui and Li, Yongqian and Luo, Yong and Zhou, Xin},
  booktitle={International Conference on Artificial Neural Networks},
  pages={56--70},
  year={2024},
  organization={Springer}
}

@article{End-to-end,
  title={End-to-end multimodal clinical depression recognition using deep neural networks: A comparative analysis},
  author={Muzammel, Muhammad and Salam, Hanan and Othmani, Alice},
  journal={Computer Methods and Programs in Biomedicine},
  volume={211},
  pages={106433},
  year={2021},
  publisher={Elsevier}
}

@article{mago,
  title={MAGO: Multi-Knowledge Aware and Global Strategy Sequence Optimizing Network for Emotional Support Conversation},
  author={Xie, Qijun and Peng, Wei},
  journal={Neurocomputing},
  volume={618},
  pages={128888},
  year={2025},
  publisher={Elsevier}
}

@inproceedings{MDDR,
  title={MDDR: Multi-modal Dual-Attention aggregation for Depression Recognition},
  author={Zhang, Wei and Zhu, En and Chen, Juan and Li, YunPeng},
  booktitle={Proceedings of the 32nd ACM International Conference on Multimedia},
  pages={321--329},
  year={2024}
}

@article{llama,
  title={Llama: Open and efficient foundation language models},
  author={Touvron, Hugo and Lavril, Thibaut and Izacard, Gautier and Martinet, Xavier and Lachaux, Marie-Anne and Lacroix, Timoth{\'e}e and Rozi{\`e}re, Baptiste and Goyal, Naman and Hambro, Eric and Azhar, Faisal and others},
  journal={arXiv preprint arXiv:2302.13971},
  year={2023}
}

@article{gpt4,
  title={Gpt-4 technical report},
  author={Achiam, Josh and Adler, Steven and Agarwal, Sandhini and Ahmad, Lama and Akkaya, Ilge and Aleman, Florencia Leoni and Almeida, Diogo and Altenschmidt, Janko and Altman, Sam and Anadkat, Shyamal and others},
  journal={arXiv preprint arXiv:2303.08774},
  year={2023}
}

@inproceedings{ResNet,
  title={Deep residual learning for image recognition},
  author={He, Kaiming and Zhang, Xiangyu and Ren, Shaoqing and Sun, Jian},
  booktitle={Proceedings of the IEEE conference on computer vision and pattern recognition},
  pages={770--778},
  year={2016}
}

@article{hubert,
  title={Hubert: Self-supervised speech representation learning by masked prediction of hidden units},
  author={Hsu, Wei-Ning and Bolte, Benjamin and Tsai, Yao-Hung Hubert and Lakhotia, Kushal and Salakhutdinov, Ruslan and Mohamed, Abdelrahman},
  journal={IEEE/ACM transactions on audio, speech, and language processing},
  volume={29},
  pages={3451--3460},
  year={2021},
  publisher={IEEE}
}

@article{OpenFace2.0,
  title={OpenFace 2.0: Facial Behavior Analysis Toolkit},
  author={ Baltrusaitis, Tadas  and  Zadeh, Amir  and  Lim, Yao Chong  and  Morency, Louis Philippe },
  journal={IEEE Computer Society},
  pages={59-66},
  year={2018},
}

@article{CMDC,
  title={Semi-structural interview-based Chinese multimodal depression corpus towards automatic preliminary screening of depressive disorders},
  author={Zou, Bochao and Han, Jiali and Wang, Yingxue and Liu, Rui and Zhao, Shenghui and Feng, Lei and Lyu, Xiangwen and Ma, Huimin},
  journal={IEEE Transactions on Affective Computing},
  volume={14},
  number={4},
  pages={2823--2838},
  year={2022},
  publisher={IEEE}
}

@inproceedings{DAIC,
  title={The distress analysis interview corpus of human and computer interviews.},
  author={Gratch, Jonathan and Artstein, Ron and Lucas, Gale M and Stratou, Giota and Scherer, Stefan and Nazarian, Angela and Wood, Rachel and Boberg, Jill and DeVault, David and Marsella, Stacy and others},
  booktitle={LREC},
  pages={3123--3128},
  year={2014},
  organization={Reykjavik}
}

@inproceedings{huang,
  title={ECR-chain: advancing generative language models to better emotion-cause reasoners through reasoning chains},
  author={Huang, Zhaopei and Zhao, Jinming and Jin, Qin},
  booktitle={Proceedings of the Thirty-Third International Joint Conference on Artificial Intelligence},
  pages={6288--6296},
  year={2024}
}

@inproceedings{HiQuE,
  title={HiQuE: Hierarchical Question Embedding Network for Multimodal Depression Detection},
  author={Jung, Juho and Kang, Chaewon and Yoon, Jeewoo and Kim, Seungbae and Han, Jinyoung},
  booktitle={Proceedings of the 33rd ACM International Conference on Information and Knowledge Management},
  pages={1049--1059},
  year={2024}
}

@inproceedings{cubemlp,
  title={CubeMLP: An MLP-based model for multimodal sentiment analysis and depression estimation},
  author={Sun, Hao and Wang, Hongyi and Liu, Jiaqing and Chen, Yen-Wei and Lin, Lanfen},
  booktitle={Proceedings of the 30th ACM international conference on multimedia},
  pages={3722--3729},
  year={2022}
}

@article{lora,
  title={Lora: Low-rank adaptation of large language models.},
  author={Hu, Edward J and Shen, Yelong and Wallis, Phillip and Allen-Zhu, Zeyuan and Li, Yuanzhi and Wang, Shean and Wang, Lu and Chen, Weizhu and others},
  journal={ICLR},
  volume={1},
  number={2},
  pages={3},
  year={2022}
}

@inproceedings{mult,
  title={MulT: Multimodal Transformer for Unaligned Multimodal Language Sequences},
  author={Zadeh, Amir and Chen, Minghai and Poria, Soujanya and Cambria, Erik and Morency, Louis-Philippe},
  booktitle={Proceedings of the 57th Annual Meeting of the Association for Computational Linguistics},
  pages={6558--6569},
  year={2019}
}

@article{iifdd,
  title={IIFDD: Intra and inter-modal fusion for depression detection with multi-modal information from Internet of Medical Things},
  author={Chen, Jian and Hu, Yuzhu and Lai, Qifeng and Wang, Wei and Chen, Junxin and Liu, Han and Srivastava, Gautam and Bashir, Ali Kashif and Hu, Xiping},
  journal={Information Fusion},
  volume={102},
  pages={102017},
  year={2024},
  publisher={Elsevier}
}

@article{gold,
  title={The Beck Depression Inventory and General Health Questionnaire as measures of depression in the general population: a validation study using the Composite International Diagnostic Interview as the gold standard},
  author={Aalto, Anna-Mari and Elovainio, Marko and Kivim{\"a}ki, Mika and Uutela, Antti and Pirkola, Sami},
  journal={Psychiatry research},
  volume={197},
  number={1-2},
  pages={163--171},
  year={2012},
  publisher={Elsevier}
}

@article{Instructerc,
  title={Instructerc: Reforming emotion recognition in conversation with a retrieval multi-task llms framework},
  author={Lei, Shanglin and Dong, Guanting and Wang, Xiaoping and Wang, Keheng and Wang, Sirui},
  journal={arXiv preprint arXiv:2309.11911},
  year={2023}
}

@inproceedings{empirical,
  title={Empirical analysis of dialogue relation extraction with large language models},
  author={Li, Guozheng and Xu, Zijie and Shang, Ziyu and Liu, Jiajun and Ji, Ke and Guo, Yikai},
  booktitle={Proceedings of the Thirty-Third International Joint Conference on Artificial Intelligence},
  pages={6359--6367},
  year={2024}
}

@article{distilling,
  title={Distilling step-by-step! outperforming larger language models with less training data and smaller model sizes},
  author={Hsieh, Cheng-Yu and Li, Chun-Liang and Yeh, Chih-Kuan and Nakhost, Hootan and Fujii, Yasuhisa and Ratner, Alexander and Krishna, Ranjay and Lee, Chen-Yu and Pfister, Tomas},
  journal={arXiv preprint arXiv:2305.02301},
  year={2023}
}

@article{multimodal,
  title={Multimodal spatiotemporal representation for automatic depression level detection},
  author={Niu, Mingyue and Tao, Jianhua and Liu, Bin and Huang, Jian and Lian, Zheng},
  journal={IEEE transactions on affective computing},
  volume={14},
  number={1},
  pages={294--307},
  year={2020},
  publisher={IEEE}
}

@article{kojima,
  title={Large language models are zero-shot reasoners},
  author={Kojima, Takeshi and Gu, Shixiang Shane and Reid, Machel and Matsuo, Yutaka and Iwasawa, Yusuke},
  journal={Advances in neural information processing systems},
  volume={35},
  pages={22199--22213},
  year={2022}
}

@article{chain,
  title={Chain-of-thought prompting elicits reasoning in large language models},
  author={Wei, Jason and Wang, Xuezhi and Schuurmans, Dale and Bosma, Maarten and Xia, Fei and Chi, Ed and Le, Quoc V and Zhou, Denny and others},
  journal={Advances in neural information processing systems},
  volume={35},
  pages={24824--24837},
  year={2022}
}

@article{T5,
  title={Exploring the limits of transfer learning with a unified text-to-text transformer},
  author={Raffel, Colin and Shazeer, Noam and Roberts, Adam and Lee, Katherine and Narang, Sharan and Matena, Michael and Zhou, Yanqi and Li, Wei and Liu, Peter J},
  journal={Journal of machine learning research},
  volume={21},
  number={140},
  pages={1--67},
  year={2020}
}

@inproceedings{tmac,
  title={Tmac: Temporal multi-modal graph learning for acoustic event classification},
  author={Liu, Meng and Liang, Ke and Hu, Dayu and Yu, Hao and Liu, Yue and Meng, Lingyuan and Tu, Wenxuan and Zhou, Sihang and Liu, Xinwang},
  booktitle={Proceedings of the 31st ACM International Conference on Multimedia},
  pages={3365--3374},
  year={2023}
}

@article{facial,
  title={A deep multiscale spatiotemporal network for assessing depression from facial dynamics},
  author={De Melo, Wheidima Carneiro and Granger, Eric and Hadid, Abdenour},
  journal={IEEE transactions on affective computing},
  volume={13},
  number={3},
  pages={1581--1592},
  year={2020},
  publisher={IEEE}
}

@article{zhaochat,
  title={Is ChatGPT equipped with emotional dialogue capabilities?},
  author={Zhao, Weixiang and Zhao, Yanyan and Lu, Xin and Wang, Shilong and Tong, Yanpeng and Qin, Bing},
  journal={arXiv preprint arXiv:2304.09582},
  year={2023}
}

@article{qwen2.5,
  title={Qwen2. 5 Technical Report},
  author={Yang, An and Yang, Baosong and Zhang, Beichen and Hui, Binyuan and Zheng, Bo and Yu, Bowen and Li, Chengyuan and Liu, Dayiheng and Huang, Fei and Wei, Haoran and others},
  journal={arXiv preprint arXiv:2412.15115},
  year={2024}
}

@article{ernie,
  title={Ernie 3.0: Large-scale knowledge enhanced pre-training for language understanding and generation},
  author={Sun, Yu and Wang, Shuohuan and Feng, Shikun and Ding, Siyu and Pang, Chao and Shang, Junyuan and Liu, Jiaxiang and Chen, Xuyi and Zhao, Yanbin and Lu, Yuxiang and others},
  journal={arXiv preprint arXiv:2107.02137},
  year={2021}
}

@article{glm,
  title={Glm-130b: An open bilingual pre-trained model},
  author={Zeng, Aohan and Liu, Xiao and Du, Zhengxiao and Wang, Zihan and Lai, Hanyu and Ding, Ming and Yang, Zhuoyi and Xu, Yifan and Zheng, Wendi and Xia, Xiao and others},
  journal={arXiv preprint arXiv:2210.02414},
  year={2022}
}

@article{attention,
  title={Attention is all you need},
  author={Vaswani, A},
  journal={Advances in Neural Information Processing Systems},
  year={2017}
}

@article{Harnessing,
  title={Harnessing multimodal approaches for depression detection using large language models and facial expressions},
  author={ Sadeghi, Misha  and  Richer, Robert  and  Egger, Bernhard  and  Schindler-Gmelch, Lena  and  Rupp, Lydia Helene  and  Rahimi, Farnaz  and  Berking, Matthias  and  Eskofier, Bjoern M. },
  journal={npj Mental Health Research},
  volume={3},
  number={1},
  year={2024},
}

@article{2023gemini,
  title={Gemini: a family of highly capable multimodal models},
  author={Team, Gemini and Anil, Rohan and Borgeaud, Sebastian and Alayrac, Jean-Baptiste and Yu, Jiahui and Soricut, Radu and Schalkwyk, Johan and Dai, Andrew M and Hauth, Anja and Millican, Katie and others},
  journal={arXiv preprint arXiv:2312.11805},
  year={2023}
}

@article{2024claude,
  title={The claude 3 model family: Opus, sonnet, haiku},
  author={Anthropic, AI},
  journal={Claude-3 Model Card},
  volume={1},
  pages={1},
  year={2024}
}

@article{Qwen2.5-VL,
  title={Qwen2.5-VL Technical Report},
  author={Bai, Shuai and Chen, Keqin and Liu, Xuejing and Wang, Jialin and Ge, Wenbin and Song, Sibo and Dang, Kai and Wang, Peng and Wang, Shijie and Tang, Jun and Zhong, Humen and Zhu, Yuanzhi and Yang, Mingkun and Li, Zhaohai and Wan, Jianqiang and Wang, Pengfei and Ding, Wei and Fu, Zheren and Xu, Yiheng and Ye, Jiabo and Zhang, Xi and Xie, Tianbao and Cheng, Zesen and Zhang, Hang and Yang, Zhibo and Xu, Haiyang and Lin, Junyang},
  journal={arXiv preprint arXiv:2502.13923},
  year={2025}
}

@article{Emotion-LLaMA,
  title={Emotion-llama: Multimodal emotion recognition and reasoning with instruction tuning},
  author={Cheng, Zebang and Cheng, Zhi-Qi and He, Jun-Yan and Wang, Kai and Lin, Yuxiang and Lian, Zheng and Peng, Xiaojiang and Hauptmann, Alexander},
  journal={Advances in Neural Information Processing Systems},
  volume={37},
  pages={110805--110853},
  year={2024}
}

@article{MMCL,
  title={Multi-Modality Collaborative Learning for Sentiment Analysis},
  author={Wang, Shanmin and Liu, Chengguang and Liu, Qingshan},
  journal={arXiv preprint arXiv:2501.12424},
  year={2025}
}

@inproceedings{BERT,
  title={Bert: Pre-training of deep bidirectional transformers for language understanding},
  author={Devlin, Jacob and Chang, Ming-Wei and Lee, Kenton and Toutanova, Kristina},
  booktitle={Proceedings of the 2019 conference of the North American chapter of the association for computational linguistics: human language technologies, volume 1 (long and short papers)},
  pages={4171--4186},
  year={2019}
}

@article{zou2024weakly,
  title={Weakly-supervised action learning in procedural task videos via process knowledge decomposition},
  author={Zou, Minghao and Zeng, Qingtian and Zhang, Xue},
  journal={IEEE Transactions on Circuits and Systems for Video Technology},
  volume={34},
  number={7},
  pages={5575--5588},
  year={2024},
  publisher={IEEE}
}

@inproceedings{liu2025prompting,
  title={Prompting Large Models for Knowledge and Reasoning Augmentation in KB-VQA},
  author={Liu, Qiang and Ying, Mengxi and Xiao, Peng and Li, Gan and Yuan, Xinpan},
  booktitle={International Conference on Intelligent Computing},
  pages={316--327},
  year={2025},
  organization={Springer}
}

@inproceedings{wang2024q,
  title={Q\&A Prompts: Discovering Rich Visual Clues through Mining Question-Answer Prompts for VQA requiring Diverse World Knowledge},
  author={Wang, Haibo and Ge, Weifeng},
  booktitle={European Conference on Computer Vision},
  pages={274--292},
  year={2024},
  organization={Springer}
}

@article{hu2025multi,
  title={A Multi-Hop Graph Reasoning Network for Knowledge-Based VQA},
  author={Hu, Zihan and You, Jiuxiang and Yang, Zhenguo and Li, Xiaoping and Xie, Haoran and Li, Qing and Liu, Wenyin},
  journal={ACM Transactions on Intelligent Systems and Technology},
  volume={16},
  number={3},
  pages={1--23},
  year={2025},
  publisher={ACM New York, NY}
}

@article{sun2022graphiqa,
  title={GraphIQA: Learning distortion graph representations for blind image quality assessment},
  author={Sun, Simeng and Yu, Tao and Xu, Jiahua and Zhou, Wei and Chen, Zhibo},
  journal={IEEE Transactions on Multimedia},
  volume={25},
  pages={2912--2925},
  year={2022},
  publisher={IEEE}
}

@article{yu2025dvlta,
  title={DVLTA-VQA: Decoupled Vision-Language Modeling with Text-Guided Adaptation for Blind Video Quality Assessment},
  author={Yu, Li and Wang, Situo and Zhou, Wei and Gabbouj, Moncef},
  journal={arXiv preprint arXiv:2504.11733},
  year={2025}
}

@article{cheng20252,
  title={S-TFCKD: Intra-Inter Set Knowledge Distillation with Time-Frequency Calibration for Speech Enhancement},
  author={Cheng, Jiaming and Liang, Ruiyu and Xu, Chao and Ni, Ye and Zhou, Wei and Schuller, Bj{\"o}rn W and Hao, Xiaoshuai},
  journal={arXiv preprint arXiv:2506.13127},
  year={2025}
}

@article{yu2025text,
  title={Text-Audio-Visual-conditioned Diffusion Model for Video Saliency Prediction},
  author={Yu, Li and Sun, Xuanzhe and Zhou, Wei and Gabbouj, Moncef},
  journal={arXiv preprint arXiv:2504.14267},
  year={2025}
}

@article{chen2022wavlm,
  title={Wavlm: Large-scale self-supervised pre-training for full stack speech processing},
  author={Chen, Sanyuan and Wang, Chengyi and Chen, Zhengyang and Wu, Yu and Liu, Shujie and Chen, Zhuo and Li, Jinyu and Kanda, Naoyuki and Yoshioka, Takuya and Xiao, Xiong and others},
  journal={IEEE Journal of Selected Topics in Signal Processing},
  volume={16},
  number={6},
  pages={1505--1518},
  year={2022},
  publisher={IEEE}
}

@article{mcfee2015librosa,
  title={librosa: Audio and music signal analysis in python.},
  author={McFee, Brian and Raffel, Colin and Liang, Dawen and Ellis, Daniel PW and McVicar, Matt and Battenberg, Eric and Nieto, Oriol},
  journal={SciPy},
  volume={2015},
  pages={18--24},
  year={2015}
}

\end{document}